\documentclass[journal,10pt]{elsarticle}

\usepackage{framed,multirow,booktabs}
\usepackage{graphicx,epsfig}
\usepackage{amssymb,amsmath,bm}
\usepackage{textcomp}
\usepackage{algpseudocode}
\usepackage{algorithm}
\usepackage{CJK}

\usepackage{amsthm}
\usepackage{lineno,hyperref}
\usepackage{xcolor}
\usepackage{cleveref}
\usepackage[section]{placeins}
\usepackage{amssymb}
\usepackage{array}
\usepackage{makecell}
\usepackage{geometry}
\usepackage{setspace} 
\journal{Journal of Pattern Recognition}

\begin{document}

\begin{frontmatter}



\title{Bidirectional Trained Tree-Structured Decoder for Handwritten Mathematical Expression Recognition}


\author{Hanbo Cheng, Chenyu Liu, Pengfei Hu, Zhenrong Zhang, Jiefeng Ma, Jun Du}


\begin{abstract}
The Handwritten Mathematical Expression Recognition (HMER) task is a critical branch in the field of OCR. Recent studies have demonstrated that incorporating bidirectional context information significantly improves the performance of HMER models. However, existing methods fail to effectively utilize bidirectional context information during the inference stage. Furthermore, current bidirectional training methods are primarily designed for string decoders and cannot adequately generalize to tree decoders, which offer superior generalization capabilities and structural analysis capacity. In order to overcome these limitations, we propose the Mirror-Flipped Symbol Layout Tree (MF-SLT) and Bidirectional Asynchronous Training (BAT) structure. Our method extends the bidirectional training strategy to the tree decoder, allowing for more effective training by leveraging bidirectional information. Additionally, we analyze the impact of the visual and linguistic perception of the HMER model separately and introduce the Shared Language Modeling (SLM) mechanism. Through the SLM, we enhance the model's robustness and generalization when dealing with visual ambiguity, particularly in scenarios with abundant training data. Our approach has been validated through extensive experiments, demonstrating its ability to achieve new state-of-the-art results on the CROHME 2014, 2016, and 2019 datasets, as well as the HME100K dataset. The code used in our experiments will be publicly available.
\end{abstract}




\end{frontmatter}
{\bf Correspondence:}
	Dr. Jun Du, National Engineering Research Center for Speech and Language Information Processing (NERC-SLIP), University of Science and Technology of China, No. 96, JinZhai Road, Hefei, Anhui P. R. China (Email: jundu@ustc.edu.cn).

\newpage

\section{Introduction}
\begin{figure}[htbp]
\centering
\hspace{1cm}
\includegraphics[width=0.8\textwidth]{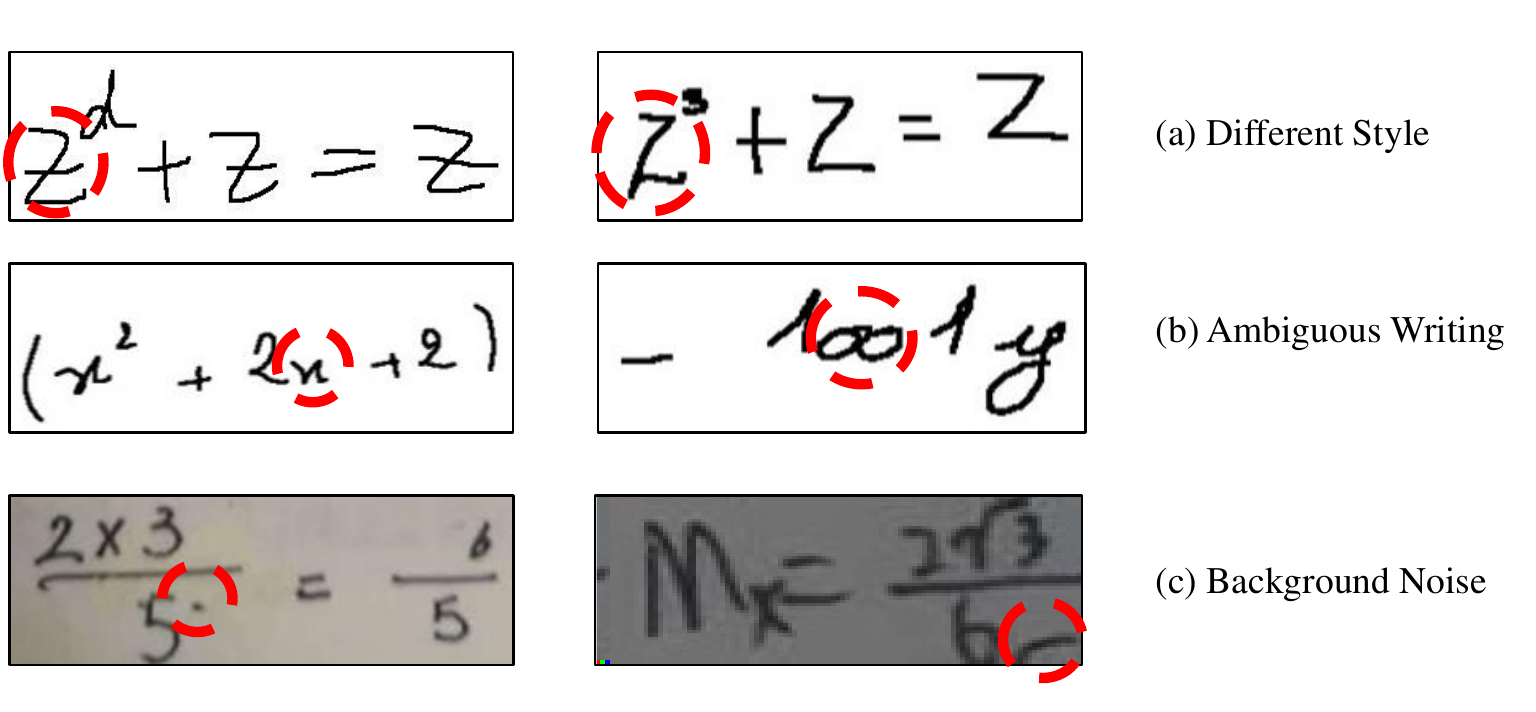}
\centering
\caption{\centering{Three categories of visual confusion. (a): The ``z" in the left and right instances have two different writing styles. (b): The ``x" on the left is similar to the ``n" and the ``0 0" on the right is similar to the ``$\infty$". (c): The background contains the noise, which is often easily recognized as ``$\cdot$". }}
\label{fig:visual_noise}
\end{figure}

Handwritten Mathematical Expression Recognition (HMER) is a significant research branch in the field of OCR. In the past decade, the importance of HMER has grown due to its wide range of applications in areas such as education and technical documentation digitization. The goal of HMER is to recognize the Mathematical Expression (ME) from a given image and convert it to LaTeX format.  Unlike normal recognition tasks such as Scene Text Recognition, the HMER requires the model to recognize individual characters while simultaneously analyzing the complex 2D structure among them. Thanks to the advancement in deep learning techniques, deep neural networks are now widely employed in HMER and have shown promising performance. However, existing HMER methods still face challenges such as low performance in dealing with misclassification caused by visual confusion \cite{SAM, Pathsig}.
As the illustration in Figure \ref{fig:visual_noise}, the misclassification resulting from the confusion of visual cues can be attributed to three specific cases in HMER: 1) Variations in writing style among different individuals for the same symbol; 2) Ambiguity or cursive writing; 3) Background noise. These visual confusions are often challenging to resolve solely through visual perception. Therefore, it is necessary to introduce more comprehensive context information \cite{BTTR, CoMER, ABM} and enhance linguistic perception \cite{Pathsig, TAP} to mitigate this issue.

The encoder-decoder architecture has significantly advanced HMER tasks, making it the mainstream method in this field. In most HMER methods, the encoder typically consists of a Convolutional Neural Network (CNN). The encoder's purpose is to extract features from the input image, providing them to the decoder which generates the target sequence. The existing HMER decoder can be divided into two categories: the String Decoder (SD) \cite{DWAP, TAP} and Tree Decoder (TD) \cite{TD, SRD}. The SD uses the LaTeX sequence as the target label, while the TD aims to generate the Symbol Layout Tree (SLT) \cite{SLT} illustrated in Figure \ref{fig:SLT} (c), which directly depicts the 2D relationship between characters. In general, the SD exhibits superior performance in learning linguistic information, leading to enhanced robustness when handling ambiguous and cursive writing \cite{TreeAug}. However, the SD struggles to directly learn the 2D spatial relationship between characters, leading to mediocre performance when handling complex ME structures \cite{TDv2}. In comparison, the TD explicitly learns the spatial relationship between characters, thus performing better in structure analysis task \cite{TD}. However, the TD has relatively weak language modeling ability and often struggles to differentiate ambiguous symbols \cite{SAN}. 

Existing methods \cite{ABM, BTTR, CoMER} have investigated the integration of bidirectional context in the HMER task. 
The method proposed by  \cite{ABM} utilizes a pair of Right-to-left (R2L) and Left-to-Right (L2R) decoders to generate the target sequence during the training stage. The output of these two decoders is constrained using the mutual learning technique \cite{DML}. Additionally, \cite{BTTR, CoMER} employ a transformer decoder to simultaneously predict the L2R and R2L LaTeX sequence.
However, these strategies are mainly tailored for string decoders and LaTeX sequence labels.
For the tree decoder, defining an inverse sequence for the Symbol Layout Tree (SLT) is challenging due to the intricate node relationships. Given the aforementioned dilemma, we design the Mirror-Flipped Symbol Layout Tree (MF-SLT) to serve as the R2L tree structure label. The MF-SLT aims to extend bidirectional training to the tree decoder. Additionally, in previous work, for the inference stage, the model still only utilizes the L2R context information \cite{ABM, BTTR}. Thus, to fully utilize bidirectional information, we propose the Bidirectional Asynchronous Training (BAT) strategy. Our proposed BAT architecture allows the model to explore and exploit the information from history and the future both in the training and inference stages, thus achieving better utilization of bidirectional context information.


Moreover, previous studies have acknowledged the significance of linguistic information in HMER task \cite{TAP, Pathsig, TreeAug, SAN}. However, the mechanism of cooperation between visual and linguistic perception, as well as their respective contributions, has not been fully evaluated.
Therefore, we separately examine the individual contributions of visual and linguistic perception using varying volumes of data in the HMER task. According to the results, we can observe that, as the training data volume grows larger (from 5k to nearly 75k), the contribution of linguistic perception increases significantly. Consequently, we conclude that enhancing linguistic perception has the potential to significantly improve the model's performance, particularly when a large amount of training data is available. Then, we propose the Shared Language Modeling (SLM) mechanism, which forces the decoder to predict the correct characters even in the absence of the visual feature. 
Our insight suggests that by reducing the model's dependence on visual features, it will emphasize learning linguistic knowledge, thereby enhancing the model's robustness in situations involving visual confusion.

The main contributions of this work are as follows: 
\begin{itemize}

    \item
    Given the lack of bidirectional context information for TD, we propose the Mirror-Flipped Symbol Layout Tree (MF-SLT) and the Bidirectional Asynchronous Training (BAT) techniques to expand the bidirectional training strategy to TD.

    \item
    We analyze the impact of both visual and linguistic information under varying volumes of training data. Then, we introduce the Shared Language Modeling (SLM) mechanism to enhance linguistic perception without introducing extra parameters. 

    \item 
    The experimental results demonstrate that our method achieves new state-of-the-art results and can be effectively generalized to both TD and SD models.
\end{itemize}




\section{Related Works}
\subsection{Handwritten Mathematical Expression Recognition}

\begin{figure}[t]
\centering
\includegraphics[width=0.95\textwidth]{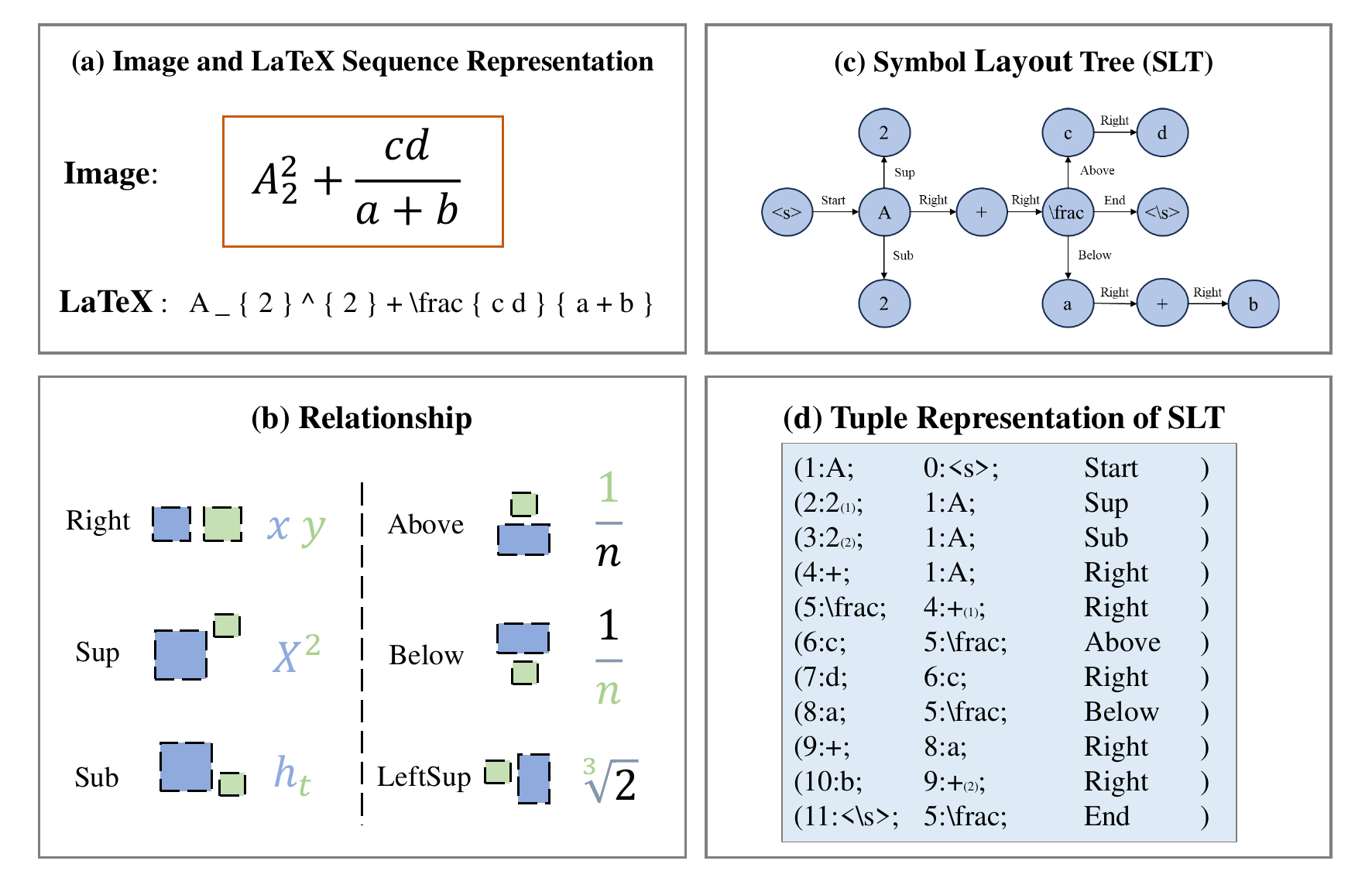}
\centering
\caption{\centering{(a) Image and LaTeX sequence representation of the Mathematical Expression (ME). (b) The categories of the relationship between symbols. (c) The Symbol Layout Tree for the ME. (d) The tuple representation of the SLT. }}
\label{fig:SLT}
\end{figure}

The HMER task consists of two sub-tasks: symbol recognition and structure analysis \cite{Chan_Yeung_2000}. In the traditional methods, the symbol recognition sub-task involves segmenting and recognizing the characters in the expression \cite{Smithies_1}. Additionally, some end-to-end methods, such as \cite{HMM} have been proposed for the symbol recognition sub-task, which helps overcome the error aggregation problem encountered in the two-stage method. Structure analysis aims to recognize the 2D spatial relationship among symbols derived from the symbol recognition stage. Previous approaches relied on human-defined syntax rules to guide the analysis of the ME's structure, known as the grammar-based method \cite{HMER_grammar_1,HMER_grammar_2}. However, designing appropriate syntax rules often presents a challenge \cite{DWAP}.

The encoder-decoder structure is the widely adopted architecture in the HMER task, which can be divided into two categories: string decoder-based methods and tree decoder-based methods. The string decoder method aims to directly generate the LaTeX label from the image, as shown in Figure \ref{fig:SLT} (a). Initially, Zhang et al. \cite{DWAP} introduced the Encoder-Decoder structure to the HMER task, utilizing a CNN-based encoder and a GRU-based decoder, which achieved remarkable performance on the CROHME dataset \cite{CROHME}. Subsequently, Zhang et al. \cite{DWAP_2} further improved the image encoder and incorporated a multi-scale feature map. To enhance the context information, Bian et al. \cite{ABM} introduced a bi-directional decoder architecture. Additionally, to improve the recognition of long sequences, Li et al. \cite{CAN} introduced a counting module into the HMER task, greatly enhancing the model's performance. Furthermore, other strategies like transformer-based decoders, such as those used in \cite{BTTR, CoMER, Transformer_decoder_HMER}, have achieved promising performance.  
The tree decoder method views the ME as a tree structure. The tree structure is widely used in the task of Handwritten Chinese Character Recognition \cite{CDF, TAN}, Document Analysis \cite{hu2022multimodal}. For HMER, the ME can be naturally represented using a tree structure \cite{TD}. Initially, \cite{SLT}  represented ME using a tree named Label Graph as illustrated in Figure \ref{fig:SLT} (b). This tree structure establishes the relationship between the nodes according to the 6 categories shown in Figure \ref{fig:SLT} (c). With the widespread adoption of encoder-decoder architectures in HMER tasks, \cite{TD} explored the feasibility of generating the tree structure label depicted in Figure \ref{fig:SLT} (d) using the encoder-decoder architecture and proposed the tree decoder. Additionally, some studies have explored the utilization of Graph Neural Networks (GNN) to predict the tree structure of HMER \cite{HMER_graph}.

\subsection{Bidirectional Training Strategy}
Bidirectional context is widely deployed in natural language processing tasks such as Neural Machine Translation (NMT) to effectively leverage context information from the history and future \cite{BiLSTM}. To further utilize bidirectional context information during the decoding stage, \cite{Asyn,Syn} respectively propose effective methods to fuse context information from the L2R and R2L directions. In OCR-related tasks, such as Scene Text Recognition, visually similar characters are prone to be misclassified by unidirectional decoder due to the lack of memory of decoding results from the future \cite{ASTER}. To tackle this problem, \cite{ASTER} has attempted to use a pair of decoders with forward and backward decoding directions.  However, for the HMER task, most methods cannot effectively utilize context information from both L2R and R2L directions, which leads to unsatisfactory robustness when decoding ambiguous writing \cite{ABM}.  Recently, the bidirectional transformer has been introduced \cite{BTTR, CoMER} for the simultaneous generation of L2R and R2L LaTeX sequences. This approach leverages comprehensive context information; however, it does not incorporate explicit supervised information for learning from the reversed direction  \cite{ABM}. A pair of bidirectional decoders are implemented in \cite{ABM}, utilizing mutual learning to enable interaction between the L2R and R2L decoders during the training stage. However, this method fails to exploit bidirectional context information during the inference stage. Furthermore, most of the bidirectional training strategies in the HMER task are specifically designed for string decoders and are incompatible with tree decoders, thus limiting their generalization.

\subsection{Language-based Recognition}
The importance of linguistic information has been extensively explored in other domains like Scene Text Recognition \cite{mishra2012scene}, where the linguistic information can effectively assist the model in addressing visual noise \cite{yu2020towards}. Numerous studies have attempted to employ linguistic information in Scene Text
Recognition by combining it with visual features or using it to rectify the recognition result \cite{NRTR, SEED}. In recent studies, powerful language models like BERT have been adopted to iteratively refine the recognition results \cite{ABI}. In the HMER task, \cite{SAN} predefined a set of syntax rules to constrain the prediction process, while \cite{SAM} exploited co-occurrence possibilities to represent the semantic information of different symbols to address the misclassification problem of visually similar characters. Furthermore, \cite{ACCV} proposed using contrastive learning to help the model learn semantic-invariant features of symbols. Some researchers have attempted to collect text-only corpora to train specialized RNN-based language models for mathematical expressions and subsequently use them to rectify prediction results \cite{TAP, AdversialLearning}. However, training extra language models can be time-consuming and requires additional datasets for language modeling. 
To alleviate this drawback, \cite{Pathsig, EMNLP_HMER} proposed using BERT as the rectification module and training BERT simultaneously with the HMER model, tremendously boosting efficiency. 
However, utilizing a specialized language model still presents the following shortcomings: 1) A substantial increase in model parameters; 2) Intrinsic involvement of a two-stage process leading to error aggregation; 3) Requiring additional training data or special strategies; and 4) Lack of flexibility to generalize to tree decoder methods.

\section{Method}

\begin{figure}[t]
\centering
\includegraphics[width=1\textwidth]{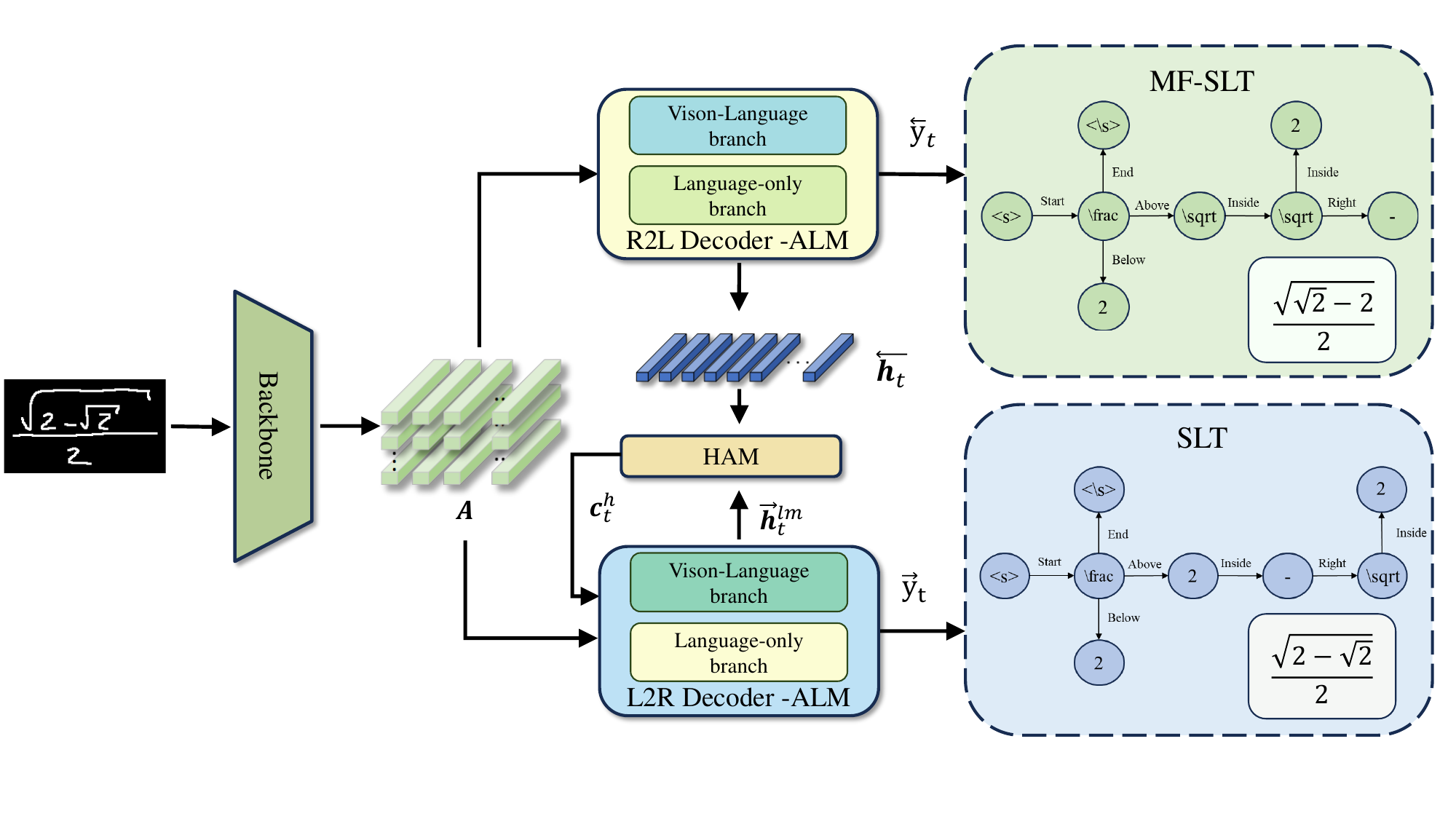}
\centering
\caption{\centering{The Bidirectional Asynchronous Training (BAT) strategy comprises a pipeline based on the encoder-decoder structure. The encoder stage generates the feature map $\boldsymbol{A}$ from the input image. The decoder stage includes a pair of R2L and L2R decoders. More specifically, the R2L decoder generates the MF-SLT, and the L2R decoder uses the hidden state produced by the R2L decoder to further predict the SLT.}}
\label{fig:pipeline}
\end{figure}
 
The pipeline of our method is depicted in Figure \ref{fig:pipeline}, which includes a backbone and a pair of tree decoders, namely the L2R decoder and R2L decoder, incorporated with the Shared Language Modeling (SLM) strategy. The Bidirectional Asynchronous Training (BAT) architecture facilitates interaction between the L2R and R2L decoders.
In the encoder stage, to ensure a fair comparison with previous methods \cite{GCN, TDv2, DWAP, ABM, CoMER, BTTR}, we employ the DenseNet \cite{DenseNet} encoder to extract the feature map $\boldsymbol{A}$ from the image.
The R2L decoder then utilizes the feature map $\boldsymbol{A}$ to predict the Mirror-Flipped Symbol Layout Tree (MF-SLT) and collect the hidden state $\overleftarrow{\boldsymbol{h}}_t$ during the decoding process, which contains both the visual and linguistic information. Once the R2L decoder completes the decoding process, the L2R decoder exploits the encoder feature map $\boldsymbol{A}$ and the hidden state from the R2L decoder to further predict the original SLT. This allows the L2R decoder access to the future decoding information collected by its R2L counterpart. In the following subsections, we will provide detailed explanations of the MF-SLT, the Bidirectional Asynchronous Training (BAT) strategy, and the Shared Language Modeling (SLM) mechanism.

\subsection{Mirror-Flipped Symbol Layout Tree}
\begin{figure}[t]
\centering
\vspace{-0.5cm}
\includegraphics[width=0.8\textwidth]{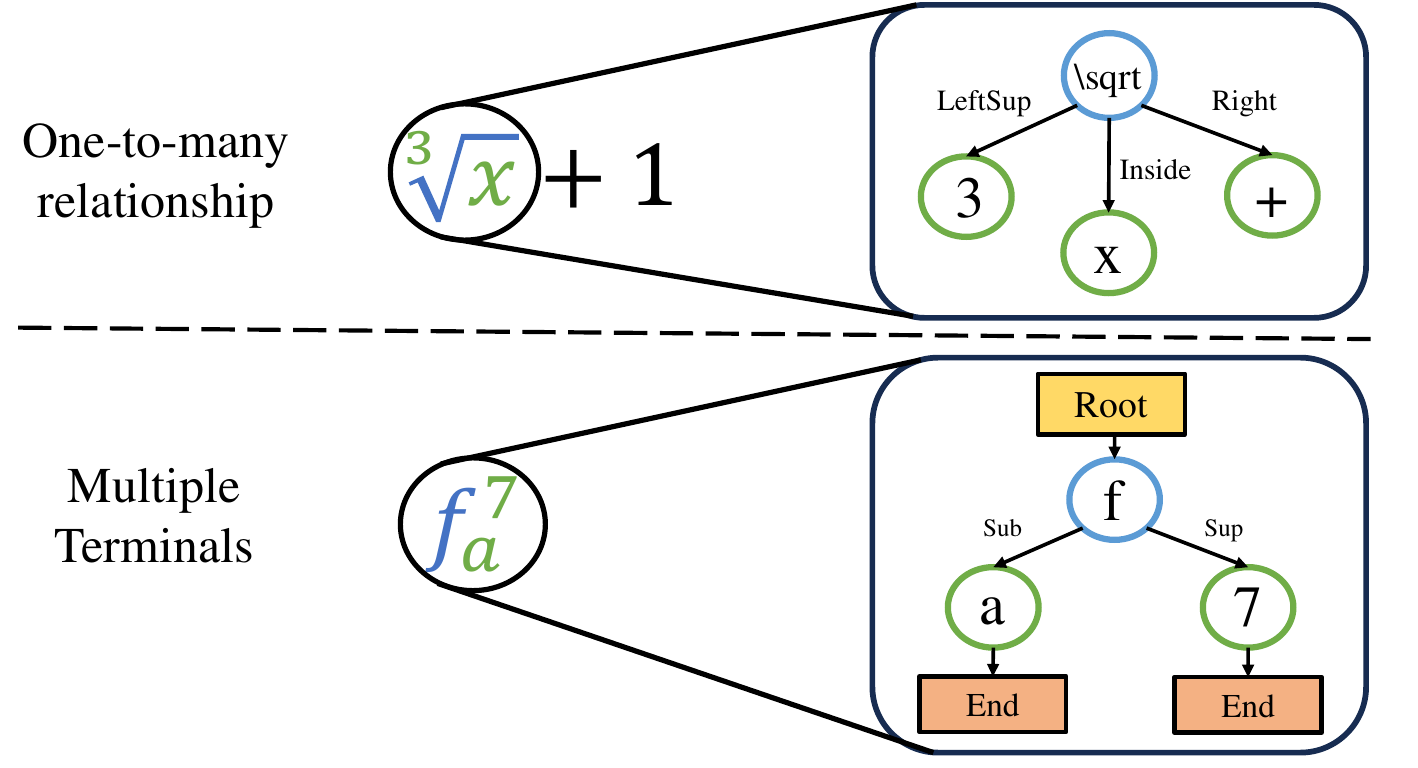}
\centering
\caption{\centering{In the tree structure labeling, the parent-child relationship is one-to-many, and there are multiple terminals. The character in blue represents the parent node, while the character in green represents the child.}}
\label{fig:SLT_prob}
\end{figure}

\begin{figure}[t]
\centering
\vspace{-0.5cm}
\includegraphics[width=0.9\textwidth]{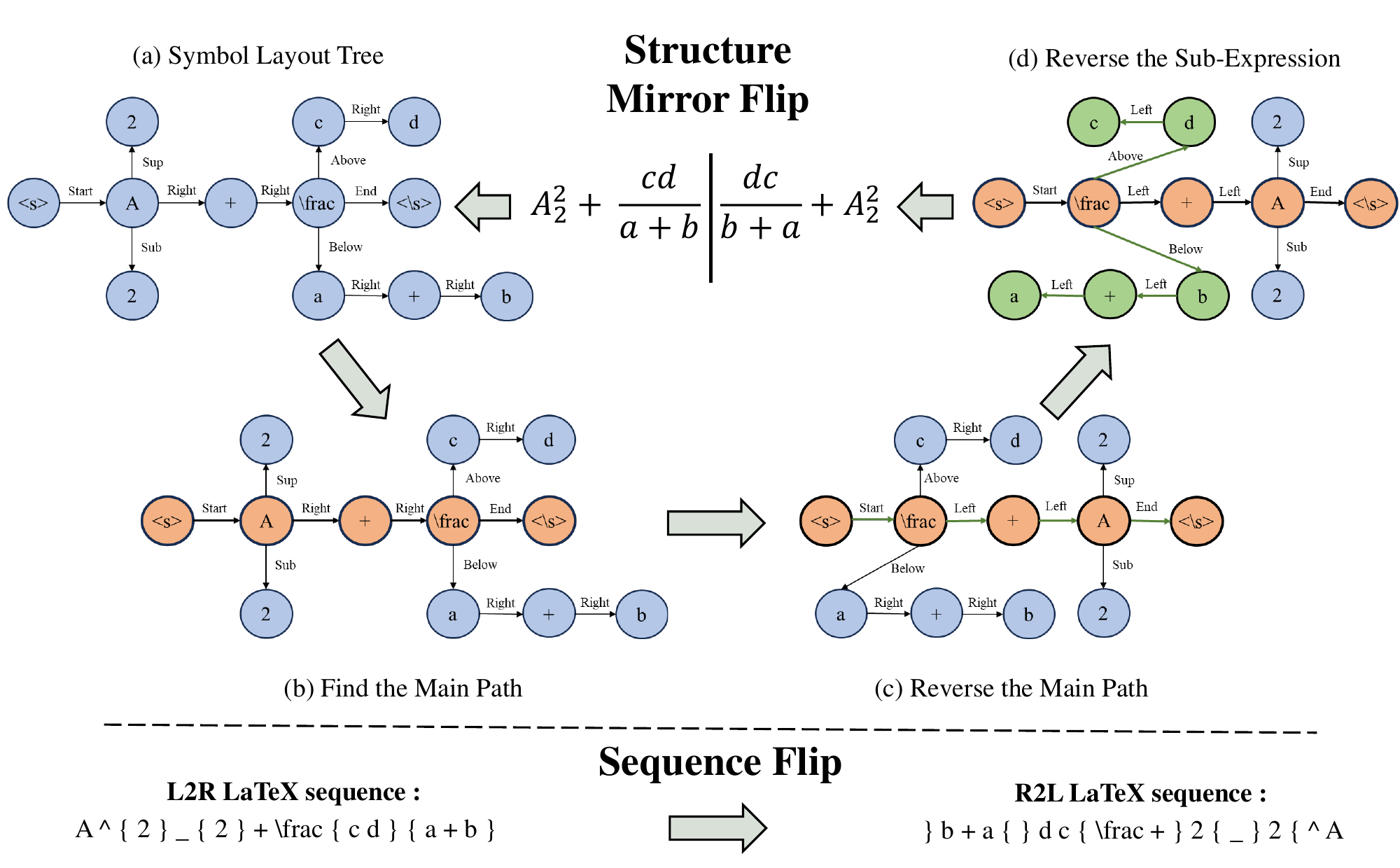}
\centering
\caption{\centering{The transformation of MF-SLT from the original SLT as well as the L2R and R2L LaTeX sequence. The relations and characters contained in ``main path" are boldened and shaded.}}
\label{fig:MFSLT}
\end{figure}

In this section, we present a novel label for the tree-decoder that guides the extraction of context information in the R2L direction. 
The LaTeX sequence label can be easily reversed to obtain the L2R and R2L labels, as demonstrated at the bottom of Figure \ref{fig:MFSLT}.
However, the situation is quite different for tree structure labels as depicted in Figure \ref{fig:SLT_prob}: 1) One-to-many problem, i.e., one parent may have more than one child; 2) There are multiple leaves that can serve as terminals. These aforementioned facts present significant difficulties in designating a new root and generating a reasonable R2L label for tree structure labels. 
However, due to the aforementioned reasons, existing bidirectional training strategies \cite{ABM, BTTR, CoMER} cannot be effectively extended to tree decoders, thereby limiting the performance of the tree decoder.

To address this issue, we propose a new SLT label called the Mirror-Flipped SLT (MF-SLT) label, which can serve as the R2L label in bidirectional training for the tree decoder. Our rationale is to emulate the way humans read an expression from right to left. We achieve this by applying a mirror-flipping operation to the 2D structure of the expression. Compared to the sequence flipping strategy utilized by \cite{BTTR, ABM}, our approach ensures that the R2L label consistently represents a valid expression, rather than abstract and human-unreadable character sequences. 

To demonstrate the generation of MF-SLT, we use the ME in Figure \ref{fig:MFSLT} as an example: 1) We define the Main Path along the ``Right” relationship, which starts from the root node; 2) We reverse the direction of the Main Path by substituting the ``Right" relationship on the main path with ``Left"; 3) We reverse the ``Right" relationship to ``Left” within the sub-tree and inverting the beginning and end of the sub-tree. In actual implement, we use the ``Forward" to represent both ``Right" and ``Left" to ensure semantic consistency for L2R and R2L labels. By the aforementioned procedure, the Main Path designates a predefined root node for the MF-SLT. Besides, we solve the one-to-many problem by reversing only the "Right" relationship. 

In comparison to sequence flipping of LaTeX sequence, the proposed MF-SLT offers a novel approach to reverse the tree structure label while preserving the accurate description of the 2D structure of MEs. By employing the MF-SLT to supervise the training of the tree decoder, the model can extract the R2L context information while keeping the semantic space of the labels unchanged compared to its L2R counterpart.

\subsection{Bidirectional Asynchronous Training Strategy }
\label{sec:BAT}

To effectively utilize bidirectional context information, we introduce a novel architecture known as Bidirectional Asynchronous Training (BAT). 
Previous studies in the HMER task \cite{ABM, BTTR} have suggested that bidirectional training enhances recognition accuracy at the end of sequences. Additionally, utilizing bidirectional context information allows the model to predict the target using information from both the past and future, thereby improving its ability to handle ambiguous symbols \cite{Asyn}. However, during the inference stage, the L2R and R2L decoders often generate results of different lengths, which presents significant challenges for alignment and fusion. As a result, existing methods struggle to effectively utilize bidirectional information during inference.

To address this issue, we present the BAT architecture, as depicted in Figure \ref{fig:pipeline}. This architecture consists of a pair of bidirectional decoders, decoding the R2L and L2R targets, respectively. 
This pair of decoders is built on the typical GRU-based HMER decoder, which can be expressed as:
\begin{gather}
	\boldsymbol{\hat{{h}}}_{t} = \text{GRU}_1(\boldsymbol{E}({\mathrm{y}}_{t-1}),  
        \boldsymbol{{h}}_{t-1}) \label{con:GRU_cells}\\ 
        \boldsymbol{{c}}_{t} = f_{{attn}}(Q = \boldsymbol{\hat{{h}}}_{t}, K = \boldsymbol{A} , V = \boldsymbol{A} ) \\
        \boldsymbol{{h}}_{t} = \text{GRU}_2(\boldsymbol{{c}}_{t},  \boldsymbol{\hat{{h}}}_{t})
\end{gather}
where $\boldsymbol{E}$ is the embedding layer. The $\text{GRU}_1$ and $\text{GRU}_2$ represent the first and second layers of GRU cells, and $\boldsymbol{\hat{{h}}}_{t}$ and $\boldsymbol{{h}}_{t}$ correspond to their respective hidden states. The function $f_{\text{attn}}$ denotes the Bahdanau Attention Mechanism \cite{addtive_attn}. The $\boldsymbol{A}$ denotes the visual features provided by the DenseNet encoder, the ${\mathrm{y}}_{t}$ represents the decoding result at the $t^{\text{th}}$ decoding step, and $\boldsymbol{{c}}_{t}$ is the visual context vector extracted from $\boldsymbol{A}$ at step $t$.

The R2L decoder predicts the MF-SLT labels $\overleftarrow{{\boldsymbol{Y}}}$ and collect the hidden states $\overleftarrow{\boldsymbol{h}}_t \in \mathbb{R}^{m}$ at each decoding step. This forms an R2L context feature map $\overleftarrow{\boldsymbol{H}} \in \mathbb{R}^{L\times m}$ that combines linguistic and visual information. The process can be expressed as:
\begin{gather}
        \text{p}(\overleftarrow{{\mathrm{y}}}_t) = \sigma( \boldsymbol{{W}}_o' \phi( \boldsymbol{{W}}_{{h}}' \overleftarrow{\boldsymbol{{h}}}_t +  \boldsymbol{{W}}_{c}' \overleftarrow{\boldsymbol{{c}}}_t + \boldsymbol{{W}}_{\mathrm{y}}' \boldsymbol{E}'({\overleftarrow{\mathrm{y}}}_{t-1})))
\end{gather}
where $ \boldsymbol{{W}}_o \in \mathbb{R}^{K\times m},  \boldsymbol{{W}}_{{h}} \in \mathbb{R}^{m\times n} ,  \boldsymbol{{W}}_{{c}} \in \mathbb{R}^{m\times D},  \boldsymbol{{W}}_{\mathrm{y}} \in \mathbb{R}^{m\times n}
,\boldsymbol{{E}} \in \mathbb{R}^{K\times m}$ are trainable parameters, and $\sigma$, $\phi$ are the softmax and maxout activation functions. 
The mark of $*'$ represents the trainable parameters in the R2L decoder.

\begin{figure}[t]
\centering

\includegraphics[width=1\textwidth]{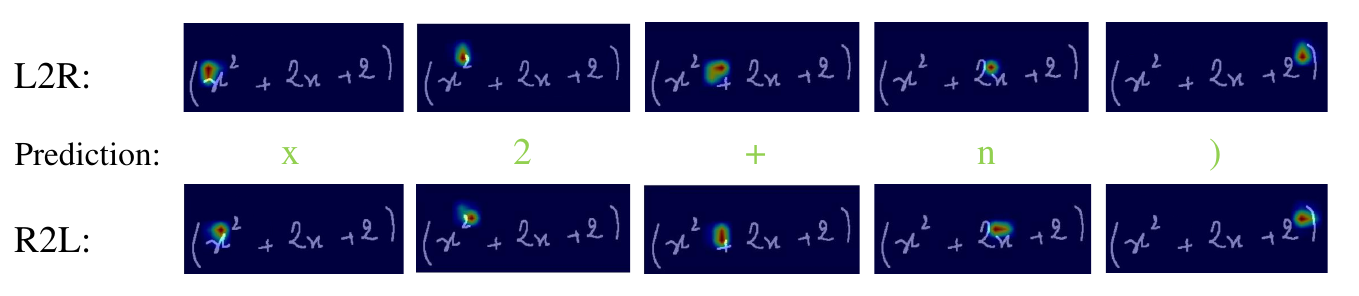}
\centering

\caption{\centering{The difference of attention distribution of L2R and R2L branches.}}
\label{fig:L2R_R2L_diff}
\end{figure}

Subsequently, the L2R decoder generates the target sequence, using the encoder output $\boldsymbol{A}$ and R2L context feature map $\overleftarrow{\boldsymbol{H}}$. To extract relevant information from the R2L context feature map $\overleftarrow{\boldsymbol{H}}$, we introduce the Hidden state Attention Module (HAM), which utilizes the Bahdanau Attention Mechanism. 
Moreover, unlike the attention module for the visual feature map $\boldsymbol{A}$ \cite{DWAP}, we use the hidden state generated by the linguistic branch in the SLM, denoted by $\overrightarrow{\boldsymbol{\hat{h}}}^{lm}_t$, as the query, which only contains the linguistic information. The mechanism of the SLM will be discussed in Section \ref{sec:section_SLM}.  The attention mechanism is employed to calculate similarities among the inputs, to capture long-range dependencies \cite{Transformer}. Since the R2L context feature map $\overleftarrow{\boldsymbol{H}}$ contains both visual and linguistic information. We observed that attention maps for $\boldsymbol{A}$ differ between the L2R and R2L branches, even though both branches generate correct predictions from features extracted by the attention map, which is illustrated in Figure \ref{fig:L2R_R2L_diff}. This observation suggests that relying on visual features for calculating similarities may lead to inaccuracies.
Consequently, we believe that incorporating discrete linguistic information into the similarity calculation will yield more accurate retrieval results. Moreover, we observed that the model’s performance is enhanced
when the linguistic information is utilized to retrieve the R2L context. To address the coverage issue, we follow the approach described in \cite{DWAP} and provide the history attention scores to the HAM, which can be denoted as:
\begin{gather}
\boldsymbol{{c}}^{hidden}_t = f_{\text{HAM}}(\overrightarrow{\boldsymbol{\hat{h}}}^{lm}_t,\overleftarrow{\boldsymbol{H}}).
\end{gather}
The L2R decoder predicts the next target through the following process:
\begin{gather}
        \text{p}(\overrightarrow{{\mathrm{y}}}_t) = \sigma( \boldsymbol{{W}}_o \phi( \boldsymbol{W}_{h} \overrightarrow{\boldsymbol{h}}_t +  \boldsymbol{{W}}_{c} \overrightarrow{\boldsymbol{c}}_t + \boldsymbol{{W}}_{c}^{hidden} \boldsymbol{{c}}^{hidden}_t +  \boldsymbol{{W}}_{\mathrm{y}} \boldsymbol{E}({\overrightarrow{\mathrm{y}}}_{t-1})))
\end{gather}
where $\boldsymbol{{W}}_o,\boldsymbol{W}_{h}, \boldsymbol{{W}}_{c}, \boldsymbol{{W}}_{c}^{hidden}, \boldsymbol{{W}}_{\mathrm{y}}$ are trainable parameters. Eventually, we optimize the recognition result from L2R and R2L branches $\overrightarrow{\hat{\boldsymbol{Y}}}, \overleftarrow{\hat{\boldsymbol{Y}}}$,  jointly with equal weight:
\begin{gather}
    Loss_{BAT} = CE(\overrightarrow{\hat{\boldsymbol{Y}}} , \boldsymbol{Y}) +  CE(\overleftarrow{\hat{\boldsymbol{Y}}} , \boldsymbol{Y})
\end{gather}

\subsection{Shared Language Modeling Mechanism}
\label{sec:section_SLM}
This section introduces the Shared Language Modeling (SLM) mechanism, which directly provides a signal to supervise the HMER decoder in learning linguistic information. The insight behind this method is that the HMER is primarily a vision-oriented task, where the model can achieve considerable performance based solely on visual perception, especially with a relatively small training set. However, an excessive dependency on visual features can compromise the model's ability to accurately classify visually similar characters. To validate this assumption, a series of experiments are conducted on the DWAP \cite{DWAP} to assess the contribution of both the visual and linguistic abilities of the model in the HMER task. The details of the experiments will be described in Section \ref{sec:motivation_exp}.

\begin{figure}[t]
\centering
\includegraphics[width=0.95\textwidth]{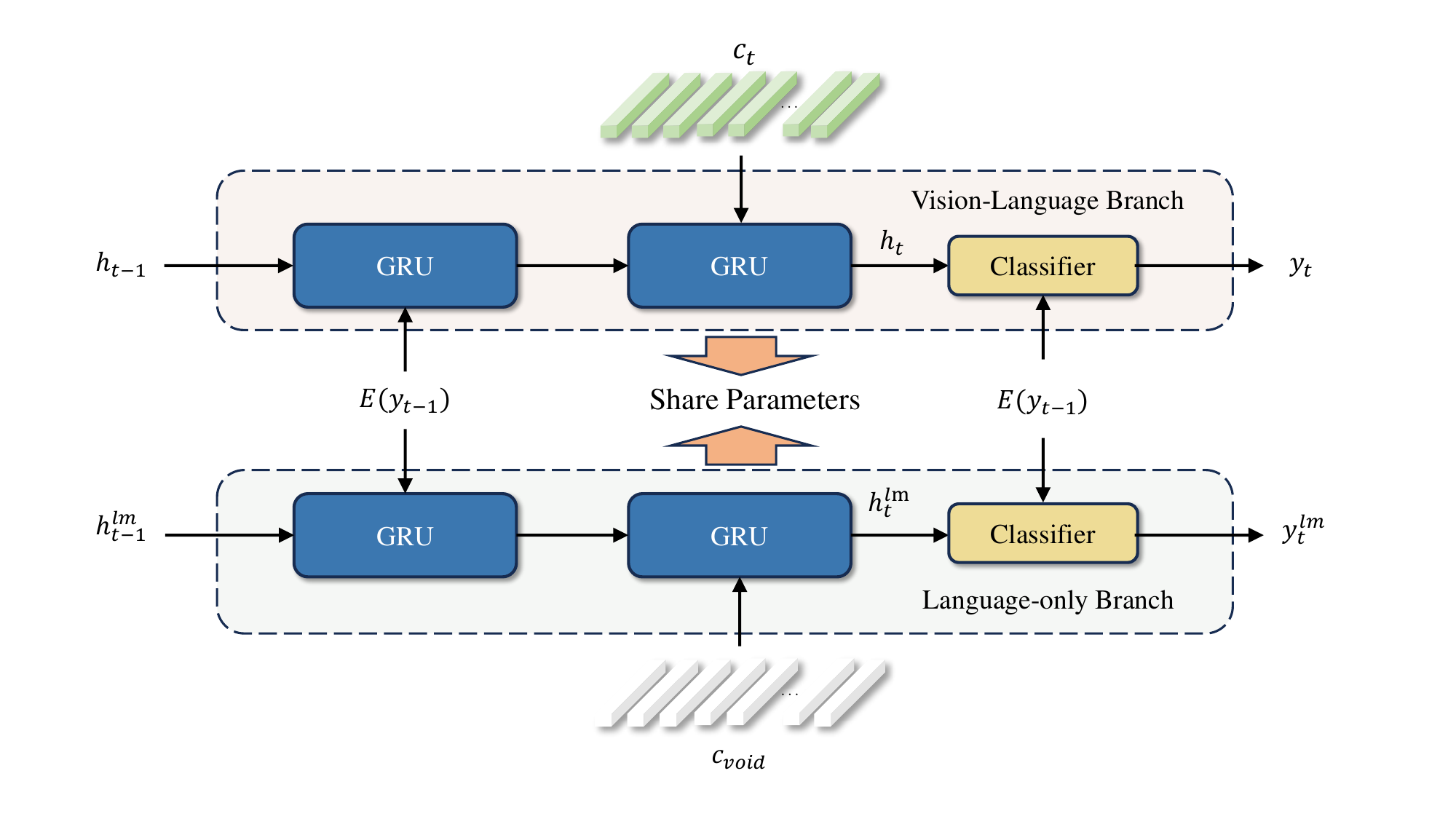}
\centering
\caption{\centering{The architecture of Shared Language Modeling (SLM) mechanism.}}
\label{fig:SLM}
\end{figure}

Based on the aforementioned belief, we propose a novel method that encourages the model to explicitly learn linguistic information along with the HMER. Illustrated in Figure \ref{fig:SLM}, the SLM decoder incorporates a dual branch structure: the vision-language branch and the language-only branch, which share parameters. The vision-language branch maintains the same structure as the regular HMER decoder as introduced in Section \ref{sec:BAT}, while the language-only branch focuses on predicting the target sequence without relying on visual features. The representation of the language-only branch is as follows:
\begin{gather}
	\boldsymbol{\hat{{h}}}^{lm}_{t} = \text{GRU}_1(\boldsymbol{E}(\mathrm{y}_{t-1}), 
        \boldsymbol{{h}}^{lm}_{t-1}) \\ 
        \boldsymbol{{h}}^{lm}_{t} = \text{GRU}_2(\boldsymbol{{c}}_{void},  \boldsymbol{\hat{{h}}}^{lm}_{t}) \\
        \text{p}({\mathrm{y}}^{lm}_t|{\mathrm{y}}_{t-1}) = \sigma( \boldsymbol{{W}}_o \phi( \boldsymbol{{W}}_{{h}} \boldsymbol{{h}}^{lm}_{t}  +  \boldsymbol{{W}}_{\mathrm{y}} \boldsymbol{E}({\mathrm{y}}_{t-1}))).
\end{gather}

The hidden states $\boldsymbol{{h}}^{lm}_{t}$ and $\boldsymbol{\hat{h}}^{lm}_{t}$ are generated by the language-only branch and exclusively carry linguistic information. The $\boldsymbol{{c}}_{void} \in \mathbb{R}^{m} $ means the zero vector with the same shape as $\boldsymbol{{c}}_t$. The loss function of the SLM can be illustrated as follows:
\begin{gather}
    L_{{SLM}} = \lambda_1 CE(\hat{\boldsymbol{Y}} , \boldsymbol{Y}) + \lambda_2 CE(\hat{\boldsymbol{Y}}^{lm} , \boldsymbol{Y}). \label{con:ALM_loss}
\end{gather}

The outputs $\hat{\boldsymbol{Y}}$ and $\hat{\boldsymbol{Y}}^{lm}$ represent the predictions made by the prime and linguistic branches of SLM, respectively. The weights assigned to these branches are denoted as $\lambda_1, \lambda_2$, and $\boldsymbol{Y}$ refers to the ground truth label.

Through the SLM mechanism, the model is encouraged to learn linguistic knowledge autonomously, reducing its dependence on visual features and improving generalization in visually ambiguous situations. Additionally, our approach avoids introducing additional parameters, and our experiment demonstrates its applicability to both TD and SD. Compared to language model-based methods \cite{TAP, Pathsig}, our method offers the following advantages: 1) We avoid the introduction of extra parameters in the inference stage; 2) Our approach is an end-to-end method, promoting better integration with the visual component and mitigating error aggregation; 3) Our method can alleviate misclassification within the same symbol category \cite{Pathsig} and can easily adapt to various datasets.

\section{Experiment}

\subsection{Dataset}

We evaluate the performance of our proposed method using two public datasets for HMER: CROHME and HME100K.

1) CROHME \cite{CROHME} is a widely used dataset in the HMER task, captured using a Digitizing Tablet. The training set of CROHME consists of 8,836 images of handwritten mathematical expressions. The CROHME is originally an online HMER dataset, we use the online trajectory points sequence to generate the offline images. The CROHME also provides three test sets: CROHME 2014, 2016, and 2019, containing 986, 1,147, and 1,199 instances, respectively. The CROHME dataset includes 101 categories of characters.

2) HME100K is a public HMER dataset proposed by \cite{SAN}. It contains a training set of 74,502 images and a testing set of 24,607 images. HME100K captures images from realistic scenes, incorporating various factors such as twists, blur, and intricate backgrounds. The dataset encompasses 245 categories of characters. In comparison to the CROHME dataset, the HME100K dataset closely aligns with real-world application scenarios and offers a significantly larger volume of data, both for training and testing. Consequently, the HME100K dataset is better suited to accurately assess the performance of the model.

\subsection{Implement Details}
We employ DenseNet as the encoder, consisting of 22 DenseBlocks. For the decoder, both the L2R and R2L decoders consist of 2 layers of GRU cells, with the hidden state dimension set to 256. In the loss function of the SLM, we set the values of $\lambda_1$ and $\lambda_2$ in Equation \ref{con:ALM_loss} to $1$ and $0.1$ respectively. 
Additionally, the weight of the L2R and R2L branches in the BAT loss is equalized. The overall loss function can be expressed as: 
\begin{gather}
    Loss = L_{SLM}'(\overrightarrow{\hat{\boldsymbol{Y}}} , \boldsymbol{Y}) +  L_{SLM}(\overleftarrow{\hat{\boldsymbol{Y}}} , \boldsymbol{Y})
\end{gather}
where the $L_{SLM}'$ and $ L_{SLM}$ are the loss of the SLM from R2L and L2R branches in the BAT.

During the training process, we set the batch size to 32 and all experiments are performed on a single NVIDIA 3090 24G GPU.
We used the Adadelta \cite{ADADELTA} as the optimizer and adopted the cosine annealing strategy to update the learning rate. The learning rate progressively increased from 0 to 2 in the first epoch and gradually decreased to 0 by the final epoch. To account for differences in data volume, we trained the model for 45 epochs on the HME100K dataset and 240 epochs on the CROHME dataset. During the inference stage, we employed the greedy search algorithm to generate the target sequence. The models' performance was evaluated using the Expression Recognition Rate (ExpRate).

\subsection{Comparison with State-Of-The-Art Methods}

In this section, we compare the performance of our method with other state-of-the-art methods in the HMER task. The results are presented in Table \ref{tab:cmp_sota_exp}. The prefix ``BAT-" denotes the integration of our proposed architecture with SLM on the baseline method. Our method, BAT-TDv2, exhibits superior performance on both the HME100K and CROHME datasets. In comparison to the baseline method, TDv2 \cite{TDv2}, our approach significantly enhances the ExpRate, achieving $5.47\%$, $5.92\%$, $2.92\%$ on CROHME 2014, 2016, 2019 respectively, and $2.22\%$ on the HME100K. While our approach falls slightly behind the GCN \cite{GCN} in performance on the CROHME 2019 dataset, we can seamlessly integrate the GCN method into our model to enhance its overall performance. It is worth noting that our method achieves the reported performance using only the greedy search strategy during the inference stage. Consequently, the ExpRate on HME100K can better demonstrate the model's performance in real application situations.

\begin{table}[ht]
 \caption{\centering{Results on the CROHME and HME100K dataset,  \textsuperscript{\dag} represents our reproduced result. The ``-lm" refers to the language model post-possessing operation. The ``SD" and ``TD" refer to the string decoder and tree decoder.}}
  \centering
  \begin{tabular}{lcccccc}
    \toprule
    \multirow{2}{*}{Model} &\multirow{2}{*}{Year} &\multirow{2}{*}{\makecell[c]{Decoder\\Type}} & \multicolumn{3}{c}{CROHME} &\multirow{2}{*}{HME100K} \\
    \cmidrule(r){4-6} 
    & & &  2014  &  2016   &  2019   \\
    \midrule
    \multicolumn{7}{c}{Method using online data} \\
    \midrule
    TAP \cite{TAP}  & 2018    & SD     & 55.37          & 50.22  & - & -\\
    SRD \cite{SRD}  & 2020     & TD    & 55.30          & 50.40  & 50.60 & -\\
    MAN \cite{MAN}  & 2021     & SD    & 54.05           & 50.56    & 52.21 & - \\
    MDR \cite{MDR}   & 2021     & TD     & 55.80          & 52.50    & 53.60 & - \\
    SCAN \cite{SCAN}  & 2021    & SD & 57.20          & 53.97    & 56.21 & - \\
    PathSig \cite{Pathsig} &2022  & SD  & 58.92  & 59.46   & 63.22 & - \\
    PathSig + lm \cite{Pathsig} &2022  & SD  & 60.34  & 59.98   & 64.22 & - \\
    \midrule
    \multicolumn{7}{c}{Method using offline data} \\
    \midrule
    DWAP\textsuperscript{\dag} \cite{DWAP}   & 2017   & SD & 50.51        & 49.34    & 48.70  & 64.70\\
        DWAP-TD \cite{TD} & 2021  & TD    & 49.10   & 48.50    & 51.40 & 62.60\\
    G2G \cite{G2G} & 2021   & SD         & 54.46          & 52.05  & -  & - \\
    ABM \cite{ABM}  &2021    & SD   & 56.85        & 52.92    & 53.96 & 65.93\\
    CAN \cite{CAN} & 2022  & SD  & 57.00       & 56.65    & 54.88 & 67.31\\
    BTTR \cite{BTTR}   & 2021  & SD  & 53.96        & 52.31    & 52.96 & 64.10\\
    TDv2 \cite{TDv2}  &  2022   & TD   & 53.56      & 55.18    & 58.72 & - \\
    SAN \cite{SAN}  & 2022    & TD   & 56.20         & 53.60     & 53.50 & 67.10\\
    CoMER\textsuperscript{\dag} \cite{CoMER} &2022  & SD & 58.57        & 57.89     & 59.71  & -\\
    GCN \cite{GCN}    & 2023  & SD  & 60.00          & 58.94    & \textbf{61.63}  & - \\
    SAM \cite{SAM} &2023  & SD  & 56.80  & 56.67   & 56.21 & 68.08\\
    \midrule
    \midrule 
    TDv2\textsuperscript{\dag} (baseline) &2022  & TD & 54.87        & 54.58    & 57.88 & 66.44 \\
    BAT-TDv2 (ours)  & 2023   & TD & \textbf{60.34}         & \textbf{60.50}     & \underline{60.80} & \textbf{68.66}\\
    \bottomrule
  \end{tabular}
  \label{tab:cmp_sota_exp}
\end{table}

\subsection{The Impact of the Linguistic Information}
\label{sec:motivation_exp}
In this section, we will provide the details of our motivation experiments of the SLM mentioned in Section \ref{sec:section_SLM}. The ability of implicit language modeling of HMER is mainly introduced by continuously providing ${\mathrm{y}}_{t-1}$, the symbol in the last decoding step. Therefore, as described in equation \ref{con:GRU_cells}, during the forward process of the GRU cells, the linguistic features primarily exist in the hidden states, namely the $\boldsymbol{\hat{h}}_{t}$ and $\boldsymbol{h}_{t}$. Furthermore, for the output of the decoder, the $ \boldsymbol{{h}}_{t}$ and ${\mathrm{y}}_{t-1}$  are directly utilized in classification:
\begin{gather}
        \text{p}({\mathrm{y}}_t|\boldsymbol{A},{\mathrm{y}}_{t-1}) = \sigma( \boldsymbol{{W}}_o \phi( \boldsymbol{{W}}_{{h}} \boldsymbol{{h}}_t +  \boldsymbol{{W}}_{c} \boldsymbol{{c}}_t +  \boldsymbol{{W}}_{\mathrm{y}} \boldsymbol{E}({\mathrm{y}}_{t-1}))).
\end{gather}

To evaluate the contribution and actual impact of linguistic information, we conducted two sets of experiments based on DWAP \cite{DWAP}. We use the following configurations: 

For configuration 1, we only change the input of the classifier by removing $\boldsymbol{{h}}_t$ and $\mathrm{y}_{t-1}$:
\begin{gather}
        \text{p}({\mathrm{y}}_t) = \sigma( \boldsymbol{{W}}_o \phi( \boldsymbol{{W}}_{c} \boldsymbol{{c}}_t ))).
\end{gather}
Specifically, in this configuration, the linguistic information still exists in the propagation process of the GRU. 

For configuration 2, in the GRU cell, we utilize a constant token $\Tilde{\mathrm{y}}$ to block the linguistic information in the GRU cells:
\begin{gather}
        \boldsymbol{\hat{{h}}}_{t} = \text{GRU}_1(\boldsymbol{E}(\Tilde{\mathrm{y}}), 
        \boldsymbol{{h}}_{t-1}).
\end{gather}

\begin{figure}[htbp]
\centering
\begin{minipage}[t]{0.48\textwidth}
\centering
\includegraphics[width=1\textwidth]{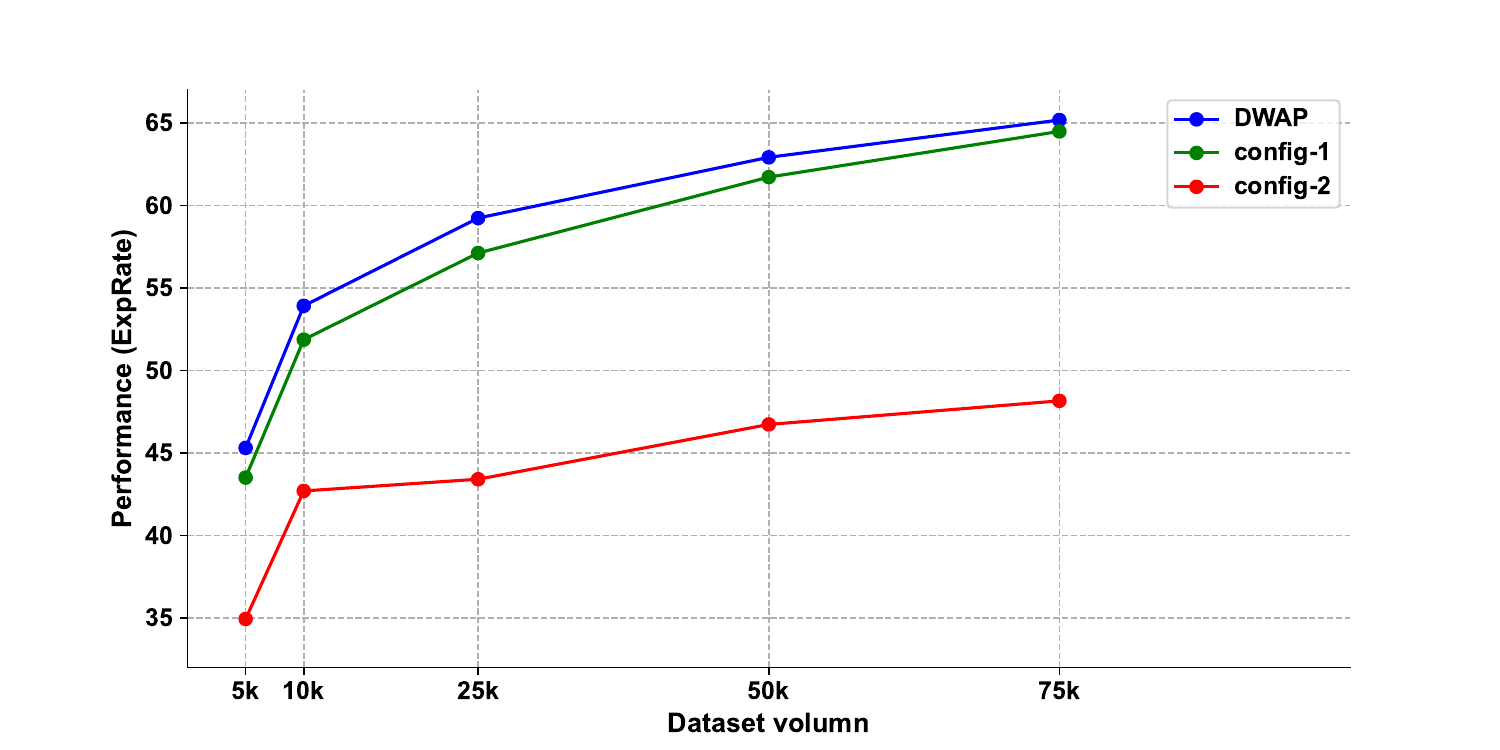}
\caption{\centering{The performance correlated with the volume of training data. }}
\label{fig:gap_ling_1}
\end{minipage}
\begin{minipage}[t]{0.48\textwidth}
\centering
\includegraphics[width=1\textwidth]{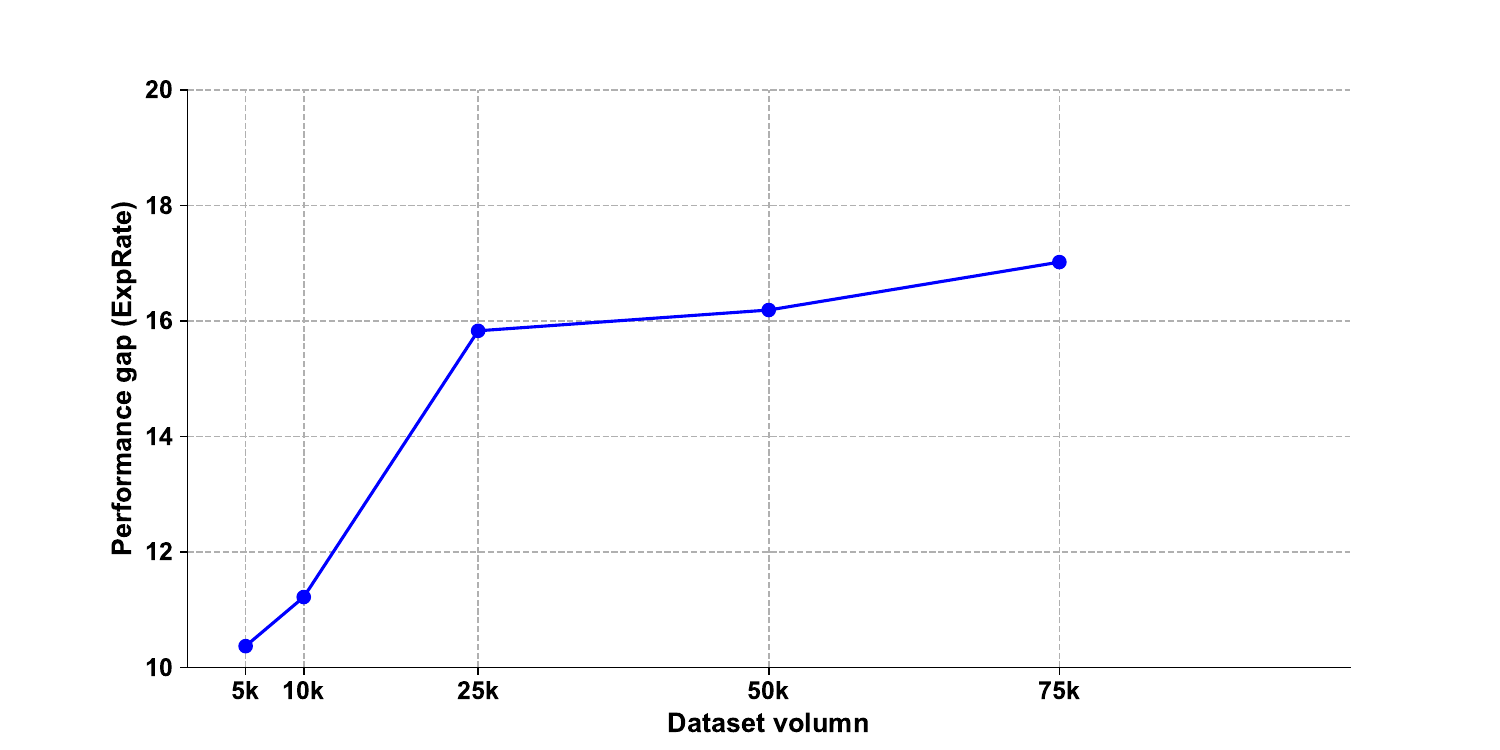}
\caption{\centering{The performance gap between the config-2 and the DWAP baseline.}}
\label{fig:gap_ling_2}
\end{minipage}
\end{figure}

The two experimental groups were trained with varying amounts of data, ranging from 5k to nearly 75k (74,502), which was randomly split from the HME100K dataset \cite{SAN}. The performance was evaluated on the test set of the HME100K. Their absolute performances are presented in Figure \ref{fig:gap_ling_1}. The performance gap between the DWAP baseline and configuration 2 is illustrated in Figure \ref{fig:gap_ling_2}, which can be viewed as the contribution of the linguistic information. The unit for the y-axis is the Expression recognition Rate (ExpRate).

\begin{figure}[t]
\centering
\vspace{-0.5cm}
\hspace{- 0.3cm}
\includegraphics[width=0.75\textwidth]{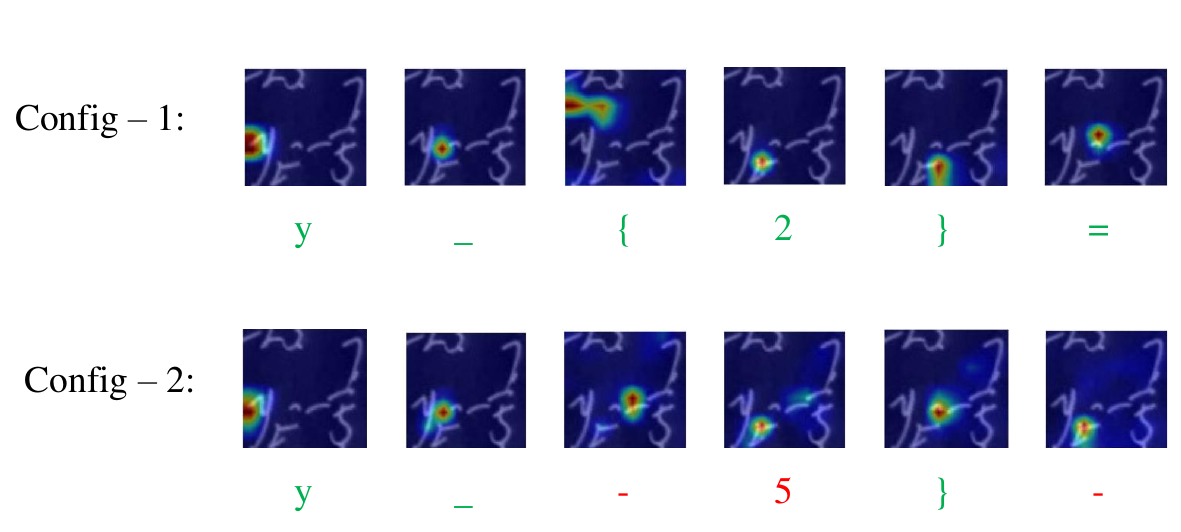}
\centering
\caption{\centering{The visualization of the attention map of configuration 1 and 2.}}
\label{fig:attn_c1_c2}
\end{figure}

In the comparison of config-1 and config-2, the linguistic information in config-2 is excluded from the query $\boldsymbol{\hat{{h}}}_{t}$ of the attention mechanism, which is responsible for retrieving relevant information from the visual feature $\boldsymbol{A}$. It is concluded that including linguistic information in the query helps the model to attend more accurately on the feature map during in each decoding step. As depicted in Figure \ref{fig:attn_c1_c2}, the absence of linguistic information in config-2 results in inferior attention accuracy. In the comparison of config-2 and the DWAP baseline, it is concluded that as the data volume increases, the impact of the linguistic information in HMER experiences a significant increase of $64.13\%$, contributing from $10.37\%$ to $17.02\%$ in performance from a data volume of 5k to almost 75k. This phenomenon suggests that when the dataset is relatively small, HMER relies more heavily on visual perception. In such cases, the model can achieve ideal performance solely relying on visual perception. However, as the size of the training set grows, linguistic information becomes increasingly important. Hence, we hypothesize that methods aiming to enhance the model’s linguistic perception will greatly improve the recognition effectiveness in HMER, especially with relatively large training sets. Without the loss of generality, we verified our assumption on the decoder by using a single-layer GRU cell \cite{CAN}, and obtained similar results.

\subsection{Ablation Study}

\subsubsection{The Effectiveness of Mirror-Flipped Symbol Layout Tree}

To assess the effectiveness of our proposed R2L label, MF-SLT, we have conducted performance tests on TDv2, which were supervised by normal SLT and MF-SLT respectively. The results are presented in Tabel \ref{tab:MFSLT}. According to the record, the performance gap between the R2L and L2R labels is relatively insignificant, indicating that the proposed MF-SLT serves as a reliable representation for the ME and enables the model to learn the R2L reading rules associated with our proposed label. Additionally, incorporating the L2R and R2L decoders significantly enhances performance, highlighting the complementarity between L2R and R2L labels.

Previous research \cite{ABM} has found that the RNN-based decoder in HMER produces high-quality outputs for prefixes but yields low-quality results for suffixes. In light of this, we evaluated the accuracy of both SLT and MF-SLT in recognizing the prefix and suffix on the CROHME, which is illustrated in Table \ref{tab:prefix_suffix}. The results indicate that our proposed MF-SLT attains higher accuracy in the suffix of the MEs, whereas normal SLT exhibits superior predictive performance in the prefix. Thus, the incorporation of the R2L branch to decode the MF-SLT has the potential to promote the recognition performance for the suffix of the target. Such a result conforms to the phenomenon found in \cite{ABM}. Besides, by using the proposed BAT architecture, we found the recognition performance of the prefix and suffix are both enhanced. Notably, the TDv2-BAT model outperforms both the single L2R and R2L TDv2 decoders in terms of the recognition rates for the prefix-5 and suffix-5.

\begin{table}[H]
 \caption{The performance of the branch supervised by SLT and MF-SLT respectively.  }
  \centering
  \begin{tabular}{lcp{1.8cm}<{\centering} cp{1.5cm}<{\centering}cp{1.5cm}<{\centering}cp{1.5cm}<{\centering}}
    \toprule
    \multirow{2}{*}{Model} 
    & \multirow{2}{*}{\makecell[c]{Label}} & \multicolumn{3}{c}{CROHME}                   \\
    \cmidrule(r){3-5}
    & & 2014          & 2016        &   2019 \\
    \midrule
    
    \multirow{2}{*}{TDv2 }  
    &  SLT (L2R)& 54.87               & 54.58      & 57.88\\
    & MF-SLT (R2L)  & 54.66               & 55.44      & 56.63\\

    \midrule
    \multirow{1}{*}{TDv2-BAT }  & MF-SLT \& SLT  & 60.34               & 60.50      & 60.80\\
    \midrule
    
  \end{tabular}
  \label{tab:MFSLT}
\end{table}

\begin{table}[H]
 \caption{\centering{The recognition accuracy of prefix and suffix of the sequence on the CROHME dataset.  The term ``prefix-n" denotes the recognition accuracy of the first n symbols in the target, while ``suffix-n" represents the recognition accuracy of the last n symbols in the target.}}
  \centering
  \begin{tabular}{lcp{1.8cm}<{\centering} cp{1.5cm}<{\centering}cp{1.5cm}<{\centering}cp{1.5cm}<{\centering}cp{1.5cm}<{\centering}}
    \toprule
    \multirow{2}{*}{Model} 
    &\multirow{2}{*}{\makecell[c]{Testing\\Set}} & \multirow{2}{*}{prefix-2} & \multirow{2}{*}{suffix-2} & \multirow{2}{*}{ prefix-5}  & \multirow{2}{*}{ suffix-5}                   \\
    \\
    \midrule
    
    \multirow{3}{*}{TDv2-R2L }  
    &  2014 & 78.70               & 81.54               & 61.99 & 64.38\\
    &  2016 & 80.12               & \textbf{83.95}      & 64.44 & 67.30\\
    & 2019  & 80.56               & \textbf{86.32}      & 64.43 & 70.38\\
    \midrule
    \multirow{3}{*}{TDv2-L2R }
    &  2014 & 86.81             & 77.38                 & 73.21 & 63.25\\
    &  2016 & 87.79             & 78.29                 & 73.48 & 62.80\\
    & 2019 & 87.91              & 79.48                 & 74.20 & 65.50\\
    \midrule
    \multirow{3}{*}{TDv2-BAT }
    &  2014 & \textbf{89.45}       & \textbf{81.64}      & \textbf{78.70} & \textbf{71.50}\\
    &  2016 & \textbf{91.54}       & 82.47             & \textbf{79.86} & \textbf{73.06}\\
    & 2019 & \textbf{89.32}        & 83.90              & \textbf{79.32} & \textbf{74.48}\\
         
    \midrule
    
  \end{tabular}
  \label{tab:prefix_suffix}
\end{table}

\subsubsection{The impact of the Bidirectional Asynchronous Training Strategy}

To evaluate the impact of different bidirectional training strategies, we conducted a comparison between our proposed Bidirectional Adaptive Training (BAT) and existing bidirectional training methods. The results of the experiments are presented in Table \ref{tab:BAT_cmp}. The baseline method, referred to as ``Uni-", represents a unidirectional approach. The term ``No interaction" refers to a model that incorporates both L2R and R2L branches but lacks any form of interaction between them. The Bidirectional Mutual Learning (BML) strategy proposed by \cite{ABM} is denoted as ``BML". Additionally, the application of the Shared Language Modeling mechanism is referred to as ``SLM", and our proposed architecture is represented by the acronym ``BAT". It is important to note that the ``BML" method was originally designed for the SD. To adapt it to the TD, we employed a mutual learning strategy between the correlated nodes in the SLT and the MF-SLT. The results demonstrate that without interaction between the L2R and R2L branches, there is no significant improvement. Although the ``BML" method improves the performance of both TD and SD to some extent, it still falls behind our proposed ``BAT" method. 

We also evaluated the enhancements in character recognition and structure analysis subtasks. During the inference stage of the model, only the ground truth sequence of characters or relationships was provided in the experiment. These results are presented in Table \ref{tab:child_relation}. The findings indicate considerable enhancements in both character recognition and structure analysis, further highlighting the superior performance of our method. Given the relatively higher accuracy of the baseline in structure analysis, our method can significantly enhance the performance of the structure analysis subtask. Additionally, the results also suggest that character recognition is the primary bottleneck of our method.

Furthermore, in our Bidirectional Adaptive Training (BAT) approach, we adopt a two-step procedure to process MEs. First, the model reads the ME from the R2L direction, collecting context information. Then, in the second pass, it reads the ME from the L2R direction, utilizing the gathered R2L context information to infer the target. To evaluate the effectiveness of the R2L context information in improving ME recognition, we conducted an additional experiment. In this experiment, we replaced the MF-SLT label in the first pass with a regular SLT. This configuration simulates a scenario where the model's task in the second pass is primarily refining the results obtained in the first stage. The results, presented in Table \ref{tab:L2RL2R}, indicate that while substituting the MF-SLT with the SLT (denoted as ``UAT-") can provide a modest performance boost, leveraging the R2L context information leads to further improvements.

\begin{table}[H]
 \caption{\centering{Compare study of different bidirectional training strategies. The ``Uni" represents the baseline method that uses a unidirectional training strategy. The ``No interaction" refers to adopting a pair of L2R and R2L branches, but no interaction is implemented. The ``BML" exhibits the method proposed by \cite{ABM}. The ``BAT" refers to our proposed bidirectional training method. The ``-SLM" represents we further implement the SLM mechanism in our method.}}
  \centering
  \begin{tabular}{lcp{1.8cm}<{\centering} cp{1.5cm}<{\centering}cp{1.5cm}<{\centering}cp{1.5cm}<{\centering}}
    \toprule
    \multirow{2}{*}{Model} 
    & \multirow{2}{*}{\makecell[c]{Bidirection\\Strategy}} & \multicolumn{3}{c}{CROHME }                   \\
    \cmidrule(r){3-5}
    & & 2014          & 2016        &   2019 \\
    \midrule
    
    \multirow{5}{*}{DWAP }  
    &  Uni & 50.51               & 49.34      & 48.70\\
    &  No Interaction           & 51.01       & 50.19      & 46.90\\
    &  BML & 54.66               & 52.39      & 51.87\\
    & BAT  & 55.33               & 55.79      & 54.12\\
    & BAT-SLM  & 55.22               & 56.15      & 55.88\\
    \midrule
    \multirow{5}{*}{TDv2 }
    &  Uni & 54.87               & 54.58      & 57.88\\
    &  No Interaction           & 55.27       & 54.57      & 57.55\\
    & BML & 56.49               & 55.45      & 58.38\\
    & BAT  & 58.01               & 58.50      & 59.71\\
    & BAT-SLM  & 60.34               & 60.50      & 60.80\\
         
    \midrule
    
  \end{tabular}
  \label{tab:BAT_cmp}
\end{table}

\begin{table}[ht]
 \caption{\centering{The experiment about the effectiveness of R2L context informaiton. The ``BAT" refers to our method extracting R2L context information in the first decoding stage but extracting L2R in the second one, and ``UAT" refers to extracting L2R context information both in the first and the second decoding stage.}}
  \centering
    \begin{tabular}{lccc}
    \toprule
    \multirow{2}{*}{Model} & \multicolumn{3}{c}{CROHME} \\
    \cmidrule(r){2-4} 
    &  2014  &  2016   &  2019   \\
    
    \midrule
    BAT-TDv2(L2R-R2L)    & \textbf{58.01}         & \textbf{58.50}     & \textbf{59.71}  \\
    UAT-TDv2(L2R-L2R)    & 56.29        & 56.58    & 59.22  \\
    \midrule
    BAT-DWAP(L2R-R2L)     & \textbf{55.33}        & \textbf{55.79}    & \textbf{54.12} \\
    UAT-DWAP(L2R-L2R)    & 54.51         & 53.87     & 53.21 \\
    \bottomrule
  \end{tabular}
  \label{tab:L2RL2R}
\end{table}

\begin{table}[ht]
 \caption{\centering{The performance gain on symbol recognition sub-task and structure parsing sub-task respectively.}}
  \centering
    \begin{tabular}{lccc}
    \toprule
    \multirow{2}{*}{Model} & \multicolumn{3}{c}{CROHME} \\
    \cmidrule(r){2-4} 
    &  2014  &  2016   &  2019   \\
    
    \midrule
    \multicolumn{4}{c}{Performance of symbol recognition sub-task} \\
    \midrule
    TDv2    & 59.03         & 62.16     & 63.97  \\
    BAT-TDv2    & 61.96(+2.93)        & 64.69(+2.53)    & 65.30(+1.23)  \\
    \midrule
    \multicolumn{4}{c}{Performance of structure parsing sub-task} \\
    \midrule
    TDv2     & 80.43         & 78.20     & 80.65 \\
    BAT-TDv2     & 82.15(+1.72)        & 81.34(+3.14)     & 84.49 (+3.84) \\
    \bottomrule
  \end{tabular}
  \label{tab:child_relation}
\end{table}

\subsubsection{The Impact of the Shared Language Modeling Mechanism}

\begin{figure}[htbp]
\centering
\begin{minipage}[t]{0.48\textwidth}
\centering
\includegraphics[width=1\textwidth]{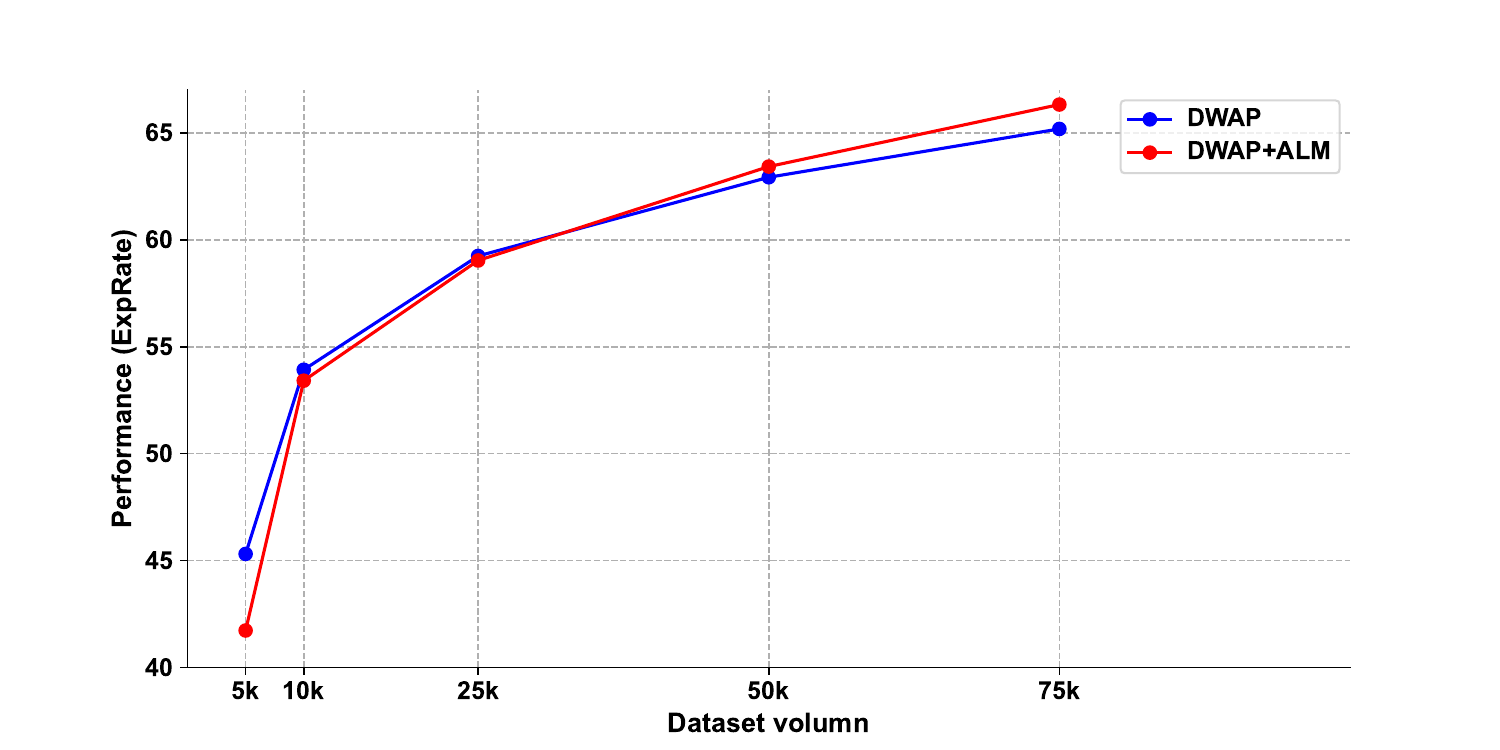}
\caption{\centering{The performance of DWAP and the DWAP with SLM correlated with data volume. }}
\label{fig:ALM_fig_1}
\end{minipage}
\begin{minipage}[t]{0.48\textwidth}
\centering
\includegraphics[width=1\textwidth]{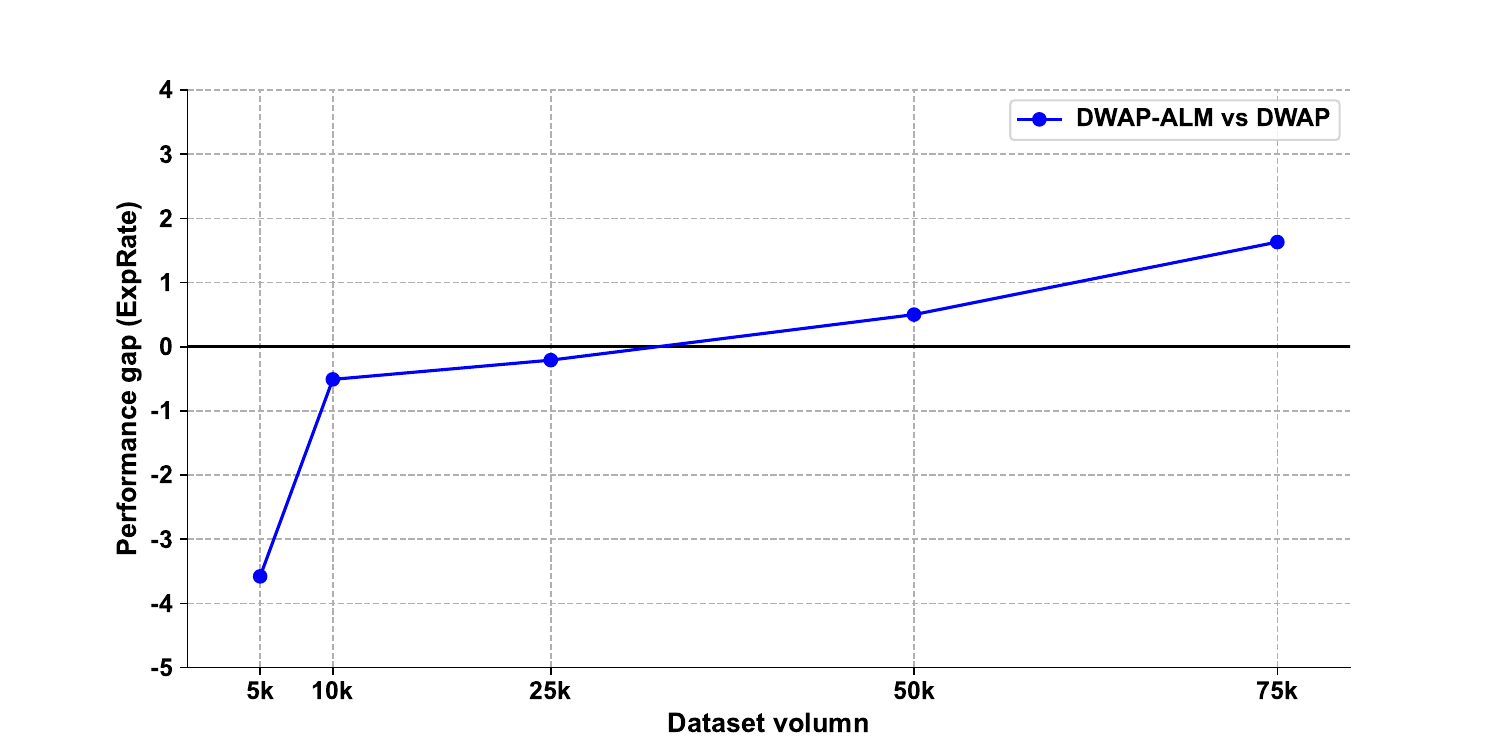}
\caption{\centering{The performance gap between DWAP and the DWAP with SLM.}}
\label{fig:ALM_fig_2}
\end{minipage}

\end{figure}

In this section, we present the contribution and properties of the Shared Language Modeling (SLM) mechanism. We first evaluated the performance improvement achieved by the SLM method on the CROHME and HME100K datasets. The results are documented in Table \ref{tab:SLM}. The SLM consistently yields a performance gain of over $1\%$ on the HME100K dataset, while the enhancement on the CROHME dataset is relatively subtle and elusive. Given the significant disparity in data volume between these two datasets, we hypothesize that this phenomenon may be attributed to the difference in data volume. To confirm our hypothesis, we further investigated the impact of training the SLM-based method on datasets with varying data volumes. The results are depicted in Figure \ref{fig:ALM_fig_1} and Figure \ref{fig:ALM_fig_2}. Specifically, Figure \ref{fig:ALM_fig_1} illustrates the absolute performance of the DWAP model with and without SLM. We gradually increased the size of the training data from 5k to nearly 75k, while assigning a weight of $0.1$ to the linguistic-only branch. On the other hand, Figure \ref{fig:ALM_fig_2} demonstrates the relative performance gap between the two models. The results indicate that, when the data volume is relatively small, the SLM model significantly lags behind the baseline. However, as the data volume increases, the performance of the SLM model gradually surpasses the baseline. This outcome suggests that linguistic information becomes increasingly significant in the presence of a large volume of training data, which further proves our hypothesis in Section 3.2.

\begin{table}[ht]
 \caption{\centering{The ablation study of Shared Language Modeling mechanism.}}
  \centering
    \begin{tabular}{lcccc}
    \toprule
    \multirow{2}{*}{Model} & \multicolumn{3}{c}{CROHME} &\multirow{2}{*}{HME100K} \\
    \cmidrule(r){2-4} 
    &  2014  &  2016   &  2019   \\
    
    \midrule
    DWAP    & 50.51        & 49.34    & 48.70  & 64.70\\
    +SLM   & 50.96(+0.45)        & 49.60(+0.26)    & 48.20(-0.50) & 66.03(+1.33) \\
    \midrule
    TDv2     & 54.87         & 54.58     & 57.88 & 66.44\\
    +SLM     & 55.98(+1.11)        & 55.10(+0.52)      & 59.22(+1.34) & 67.82(+1.38)\\
    \bottomrule
  \end{tabular}
  \label{tab:SLM}
\end{table}

\begin{CJK}{UTF8}{gkai}	
\subsection{Case Study}

\begin{figure}[t]
\centering
\includegraphics[width=1\textwidth]{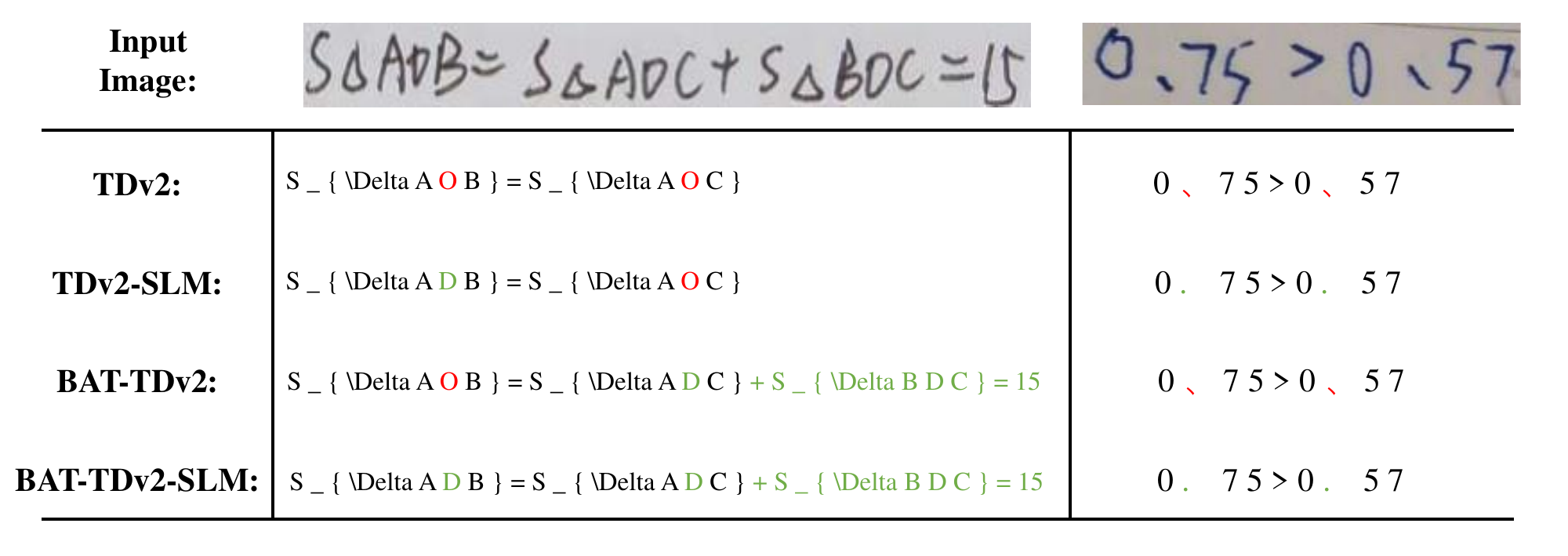}
\centering
\caption{\centering{Example for the rectification of Bidirectional Asynchronous Training (BAT) and Shared Language Modeling (SLM).   }}
\label{fig:case_study}
\end{figure}

In this section, we demonstrate how our method rectifies different types of recognition errors by several examples. As a baseline method, we choose TDv2 and compare its performance with our proposed BAT and SLM, as well as their cooperation. To illustrate the effectiveness of our method, we provide two examples, as shown in Figure \ref{fig:case_study}. The first example highlights the omission of the left side of the sequence when models do not adopt the BAT strategy. Additionally, the SLM further improves the utilization of bidirectional context information and rectifies the recognition error of the ambiguous ``D" in the image. In the second example, models without the SLM module fail to differentiate between the ``." and the ``、". However, the model equipped with the SLM module can easily recognize that the ME represents an inequality, thus realizing that the ``." is more likely to appear in such a situation.

\begin{figure}[t]
\centering
\includegraphics[width=1\textwidth]{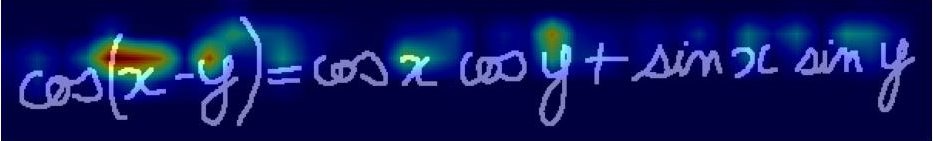}
\centering
\caption{\centering{The R2L context information, which the L2R branch attends to during the decoding step of first ``x" (the left third symbol).   }}
\label{fig:case_study_2}
\end{figure}

Additionally, we present a visualization of the attention score for the R2L hidden state in Figure \ref{fig:case_study_2}. Based on the example, when the L2R decoder recognizes the first ``x" in the sequence, the HAM module not only attends to the corresponding decoding step in the R2L branch but also considers other related characters such as ``x" and ``y". It is worth noticing that these related symbols are not accessible in the original L2R decoder. By incorporating R2L context information, the model leverages future prediction to enhance the prediction.

\end{CJK}

\section{Conclusion}

In this paper, we incorporate bidirectional context information into the tree decoder for the HMER task. To extract R2L context information by the tree decoder, we introduce a novel tree structure label called Mirro Flipped Symbol Layout Tree (MF-SLT). To fully exploit bidirectional context information in both the training and inference stages, we propose a new training architecture called Bidirectional Asynchronous Training (BAT). Additionally, we introduce the Shared Language Modeling (SLM) mechanism to enhance the model's ability to learn linguistic information and address misclassifications caused by visual noise. Through extensive experiments, we demonstrate the effectiveness and cooperation of BAT and SLM. Furthermore, the proposed architecture can be adapted to other string decoders and improve their performance. The experiment results show that our method significantly improves the recognition performance of the baseline and achieves state-of-the-art results.









\bibliography{BAT}
\end{document}